\documentclass[runningheads]{llncs}
\usepackage[nolist]{acronym}
\usepackage{graphicx}
\usepackage{booktabs}
\usepackage{xcolor}
\usepackage{subcaption}
\usepackage{microtype}
\usepackage[pagebackref=true,breaklinks=true]{hyperref}
\hypersetup{
     colorlinks,
     linkcolor={red!75!black},
     citecolor={blue!75!black},
     urlcolor={blue!75!black},
}

\usepackage[english,capitalise,noabbrev]{cleveref}
\usepackage{color}

\newcommand{\repeatthanks}{\textsuperscript{\thefootnote}}

\begin{document}
\begin{acronym}
    \acro{cer}[CER]{Character Error Rate}
    \acro{cnn}[CNN]{Convolutional Neural Network}
    \acro{htr}[HTR]{Handwritten Text Recognition}
    \acro{ocr}[OCR]{Optical Character Recognition}
    \acro{rnn}[RNN]{Recurrent Neural Network}
    \acro{cnn}[CNN]{Convolutional Neural Network}
    \acro{ctc}[CTC]{Connectionist Temporal Classification}
    \acro{crnn}[CRNN]{Convolutional Recurrent Neural Network}
    \acro{lstm}[LSTM]{Long Short-Term Memory}
    \acro{s2s}[S2S]{sequence-to-sequence}
\end{acronym}

\title{Combining OCR Models for Reading Early Modern Printed Books\thanks{Supported by the Deutsche Forschungsgemeinschaft (DFG) - Project number 460605811.}}
\author{Mathias Seuret\inst{1}\thanks{Both authors contributed equally to this research.}\orcidID{0000-0001-9153-1031} \and
Janne van der Loop\inst{2}\repeatthanks\orcidID{0000-0001-7486-669X} \and
Nikolaus Weichselbaumer\inst{2}\orcidID{0000-0002-8065-0390} \and
Martin Mayr\inst{1}\orcidID{0000-0002-3706-285X} \and
Janina Molnar\inst{2}\orcidID{0000-0003-4729-0933} \and
Tatjana Hass\inst{2}\orcidID{0000-0002-1254-3403} \and
Florian Kordon\inst{1}\orcidID{0000-0003-1240-5809} \and
Anguelos Nicolau\inst{3}\orcidID{0000-0003-3818-8718} \and
Vincent Christlein\inst{1}\orcidID{0000-0003-0455-3799}}
\authorrunning{Seuret et al.}
\institute{Friedrich-Alexander-Universit\"at Erlangen-N\"urnberg\\ \email{first.lastname@fau.de}\\
\and
University of Mainz \\
\email{\{jannevanderloop,weichsel\}@uni-mainz.de}
\and
University of Graz \\
\email{anguelos.nicolaou@gmail.com}
}
\maketitle              %

\begin{abstract}
In this paper, we investigate the usage of fine-grained font recognition on OCR for books printed from the 15th to the 18th century.
We used a newly created dataset for OCR of early printed books for which fonts are labeled with bounding boxes.
We know not only the font group used for each character, but the locations of font changes as well.
In books of this period, we frequently find font group changes mid-line or even mid-word that indicate changes in language.
We consider 8 different font groups present in our corpus and investigate 13 different subsets: the whole dataset and text lines with a single font, multiple fonts, Roman fonts, Gothic fonts, and each of the considered fonts, respectively.
We show that OCR performance is strongly impacted by font style and that selecting fine-tuned models with font group recognition has a very positive impact on the results.
Moreover, we developed a system using local font group recognition in order to combine the output of multiple font recognition models, and show that while slower, this approach performs better not only on text lines composed of multiple fonts but on the ones containing a single font only as well.

\end{abstract}

\section{Introduction}
Libraries and archives are publishing tremendous amounts of high-quality scans of documents of all types.
This leads to the need for efficient methods for dealing with such an amount of data~--~in many cases, having the documents' content as a text file instead of an image file can be of great help for scholars, as it allows them to look for specific information easily.

In this work, we focus on \ac{ocr} early modern prints~--~documents printed between the 15th and the 18th centuries~--~currently held by German libraries.
Early print frequently used a large range of font styles, much more than today, where most text, such as in this paper, is displayed or printed in some variety of Antiqua.
Our investigations are based on the prior assumption that the appearance, or shapes, of fonts, has an impact on the performance of \ac{ocr} models.
In order to evaluate the impact of fonts on \ac{ocr} and whether they can be used for improving \ac{ocr} results, we couple fine-grained recognition and \ac{ocr}.
We do not try to differentiate individual fonts -- like a particular size of Unger Fraktur -- since this information would be too granular and of limited use for \ac{ocr}. 
It would also be very challenging to label at a large scale, especially considering that early printers typically had their own, handcrafted fonts.
Instead, we differentiate eight font groups -- Antiqua, Bastarda, Fraktur, Gotico-Antiqua, Italic, Rotunda, Schwabacher and Textura. 
These font groups contain large numbers of fonts that share certain stylistic features but may have different sizes, proportions or other individual design peculiarities. 

The two main focuses of this paper are the impact of font groups on \ac{ocr} accuracy and the combination of font group classification and \ac{ocr} methods.
In particular, we (1)~evaluate the impact of selecting models fine-tuned on specific font-groups on a line basis, (2)~evaluate the impact of splitting text lines according to a fine-grained font-group classifier, and (3)~develop a novel combined OCR model denoted as COCR,\footnote{\url{https://github.com/seuretm/combined-ocr}} which outperforms the other systems in case of multiple fonts appearing in one line.\footnote{On an intermediate version of EMoFoG (Early Modern Font Groups), DOI 10.5281/zenodo.7880739, to appear.}

\section{State of the Art}

\paragraph{Optical Character Recognition}

The field of \acf{ocr} and \acf{htr} has seen remarkable progress as a result of the \acf{ctc} decoding, introduced by Graves et al.~\cite{graves2006connectionist}.
Subsequent research, such as Puigcerver et al.~\cite{puigcerver2017are}, combined the \ac{ctc} loss function with the hybrid architecture of \acp{crnn}. 
Bluche et al.~\cite{bluche2017gated} further enhanced this architecture by incorporating gated mechanisms within the convolutional layers to emphasize relevant text features. 
Michael et al.~\cite{michael2019evaluating} followed an alternative direction by employing \acf{s2s} models for \ac{htr}. These approaches consist of an encoder-decoder structure and often utilize a \ac{cnn} with a 1-D bidirectional \acf{lstm} network as the encoder and a vanilla \ac{lstm} network as the decoder, producing one character at a time. The focus of their work lies in the comparison of the attention mechanism linking the encoder and the decoder.
Yousef et al.~\cite{yousef2020accurate} and Coquenet et al.~\cite{coquenet2020recurrence,coquenet2021span,coquenet2022vertical} achieved state-of-the-art results in \ac{htr} on benchmark, such as IAM database~\cite{marti2002the} and RIMES~\cite{grosicki2011icdar}, by using recurrence-free approaches.
Kang et al.~\cite{kang2022pay} improved sequential processing of text lines by substituting all \ac{lstm} layers with Transformer layers~\cite{vaswani2017attention}. 
Wick et al.~\cite{wick2021transformer} extended this concept by fusing the outputs of a forward-reading transformer decoder with a backward-reading decoder.
Diaz et al.~\cite{diaz2021rethinking} examined various architectures with their proposed image chunking process, achieving the best results with a Transformer-CTC-based model in conjunction with a Transformer language model. 
Li et al.~\cite{li2021trocr} accomplished impressive results on the IAM database through the use of 558 million parameters \ac{s2s}-Transformer-based model without convolutional layers, as well as through massive pre-training utilizing 684 million text lines extracted from PDF documents and 17 million handwritten text lines. 
In their study of MaskOCR, Lyu et al.~\cite{lyu2022maskocr} demonstrated the benefits of self-supervised pre-training for \ac{ocr} on Chinese documents and scene text recognition. For pre-training, the model reconstructed masked-out patches of the text line images using a thin decoder. For the downstream task, the reconstruction decoder was discarded, and the remaining parts of the model underwent fine-tuning.
To improve the integration of \ac{ctc}-predictions and decoder outputs, Wick et al.~\cite{wick2022rescoring} calculated a CTC-prefix-score to enhance the sequence decoding performance of both the transformer decoder and the language model.
An ensemble of $N$ models is used by Reul et al.~\cite{reul2018improving} in a voting strategy.
They split their training data into $N$ subsets. Each of these subsets is used as validation data for a single model trained on the remaining $N-1$ other subsets. Inference is done by averaging the results of all models of the ensemble.
Wick et al.~\cite{wick2021one} refine this approach~--~ instead of training the models independently, they use a masking approach allowing to combine their losses and thus train them together.

\paragraph{Font Type Recognition}
This task has been around for a very long time, with, for example, some publications on this topic in 1991 at the first ICDAR conference~\cite{fossey1991100}.
Each font is typically considered as being one class. We however are dividing the fonts into different font groups.
In~\cite{nicolaou2014local}, the authors produce descriptors based on local binary patterns~\cite{ojala2002multiresolution} and compare them with the nearest neighbor algorithm to classify 100 fonts and sizes for Arabic text.
In~\cite{chen2021henet}, the authors use a ResNet~\cite{he2016deep} and maximum suppression to identify 1,116 different fonts in born-digital images with high accuracy.
A larger dataset with over 3300 fonts is used in~\cite{yang2019hanfont}.
As many of the fonts are visually practically indistinguishable, the authors create groups of fonts by clustering them and evaluate several \ac{cnn} architectures for either font or cluster classification.
In~\cite{seuret2019dataset}, fonts from historical books are classified, but the proposed system is restricted to outputting only one class per page.
This is done by average voting on patches, using a DenseNet~\cite{huang2017densely}.
In~\cite{tensmeyer2017convolutional}, the authors use a ResNet pre-trained on synthetic data both for script and font classification.
They do so by combining the scores on randomly selected patches that have been filtered to reject the ones containing only background.

\paragraph{Novelty of this Work}
It differs from previous publications in two main points.
First, while we also investigate the use of ensembles of \ac{ocr} models, we combine their outputs using a locally weighted average using font classification, which has not been done in other works.
Second, we push the boundary of the font group classification granularity as we aim to find where font groups change across a single text line.
Another difference, although of lower importance, is that we consider the font group classification as an auxiliary task for \ac{ocr}, and thus use \ac{ocr} evaluation as a proxy for investigating the usefulness of font group classification.

\section{Method\textbf{s}}

\subsection{One Model to OCR Them All: Baseline}
This baseline approach consists of training a single \ac{ocr} model on the whole dataset, regardless of any font information.
We based our architecture, detailed in \cref{tab:architecture}, on the \ac{ctc}-based model by Wick et al.~\cite{wick2022rescoring}. However, we employed a finer scale in the first convolutional layer of the feature extractor, a \ac{cnn}~\cite{fukushima1979neural}, than their original model and halved the kernel and stride sizes. We further decreased the number of neurons in the recurrent layers to 128, for two reasons:
First, we make the assumption that \ac{ocr} does not require an \ac{rnn} as powerful as \ac{htr}, and, second, one of the systems which makes use of multiple instances of the baseline is memory-greedy.

\begin{table}[t]
\caption{Base model architecture}\label{tab:architecture}
\begin{tabular}{lcrccl}
\toprule
Operation & Kernel size & Outputs & Stride & Padding & Comment\\
\midrule
Convolution  & $3\times 2$ &             8~~ & $2\times 1$ &          -- &         then ReLU \\
Convolution  & $6\times 4$ &            32~~ & $1\times 1$ & $3\times 1$ &         then ReLU \\
Max pooling  & $4\times 2$ &            --~~ & $4\times 2$ &          -- &                -- \\
Convolution  & $3\times 3$ &            64~~ & $1\times 1$ & $1\times 1$ &         then ReLU \\
Max pooling  & $1\times 2$ &            --~~ & $1\times 2$ &          -- &         then ReLU \\
Mean (vertical axis) & --  &            --~~ &          -- &          -- &                -- \\
Linear       &         --  &           128~~ &          -- &          -- &                -- \\
Bi-directional LSTM~\cite{hochreiter1997long}  & --  & $2\times 128$~~ &          -- &          -- &    3 layers, tanh \\
Linear       &         --  &           153~~ &          -- &          -- & 152 chars + blank \\
\bottomrule 
\end{tabular}
\end{table}

\subsection{Font-Group-Specific OCR Models}
\label{ssct:fgs}
We produce font-group-specific \ac{ocr} models by fine-tuning the baseline.
Training them from scratch, instead of using the baseline as starting point, led to high \ac{cer}.
Note that it is required to know the font group of text lines in order to apply the corresponding model.

\subsection{Select One Model per Line: SelOCR}
\begin{figure}[tb]
    \begin{center}
        \includegraphics[width=\textwidth]{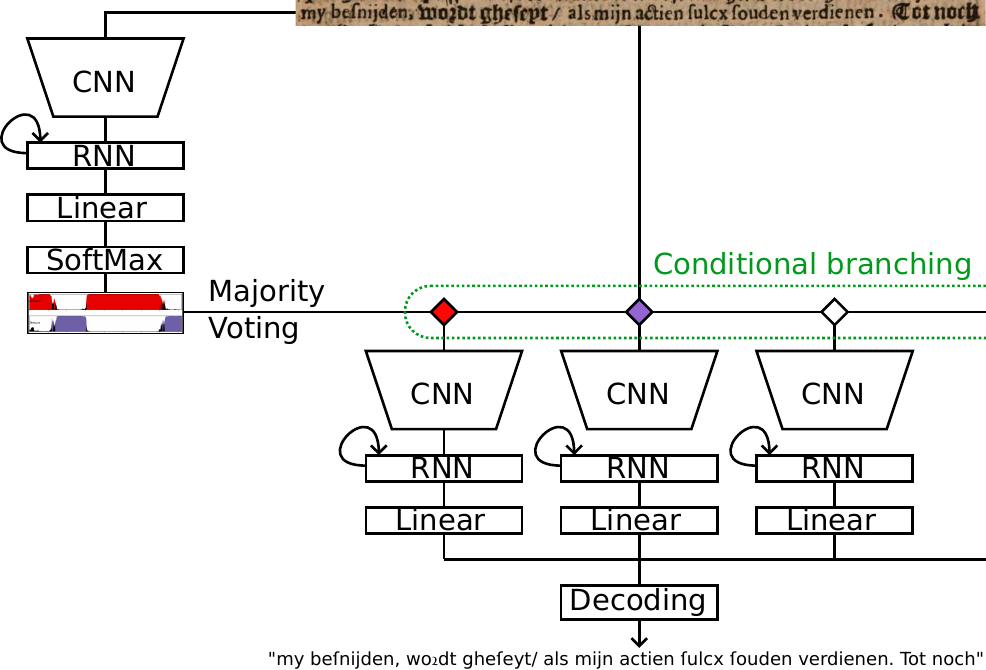}
    \end{center}
    \caption{
        Pipeline of the SelOCR system.
    }
    \label{fig:selocr}
\end{figure}

Having font-group-specific models is nice, but of little use unless we know which font group is present in the data to process.
For this reason, we developed the Selective OCR approach: a network first recognizes the font group of the input text line, and then the corresponding \ac{ocr} model is applied.
This process is illustrated in \cref{fig:selocr}.

\subsection{Splitting Lines in Heterogeneous Parts: SplitOCR}
\label{ssct:splitocr}
As font groups can change within a text line (even in the middle of a word), we can try to split text lines into homogeneous sequences of a single font group, and process each of them with an adequate \ac{ocr} model.

We first use a classifier to estimate the font group of every pixel column of the text line to process.
Then, any homogeneous segment of a width smaller than the height of the text line is split into two parts, and each gets merged with the adjacent segment.
This leads to heterogeneous segments where majority voting is applied to retain only a single class.
As the last post-processing step of the classification result, adjacent segments with the same class are merged.

Each segment is then processed with the \ac{ocr} model corresponding to the majority class of the segments' pixel columns.

\subsection{Combining OCR Models: COCR}
\ac{rnn} benefit much from processing longer sequences instead of short, independent inputs.
Thus, splitting text lines, as it is done with the SplitOCR approach, might be sub-optimal, as \ac{ocr} models might not see the whole input and benefit from advantages of processing a sequence, such as language models implicitly learned by \ac{rnn}.

For this reason, we developed one more approach based on the combination of the outputs of \ac{ocr} models using font group classification.
Font group classification is done with a model having a similar structure as the \ac{ocr} models, but with only 32 neurons in a single recurrent layer, as many outputs as there are font groups, and a softmax activation function after the output layer.
Thus, the classifier and the \ac{ocr} models all have the same output sequence length, and can easily be merged.

\begin{figure}[tb]
    \begin{center}
        \includegraphics[width=\textwidth]{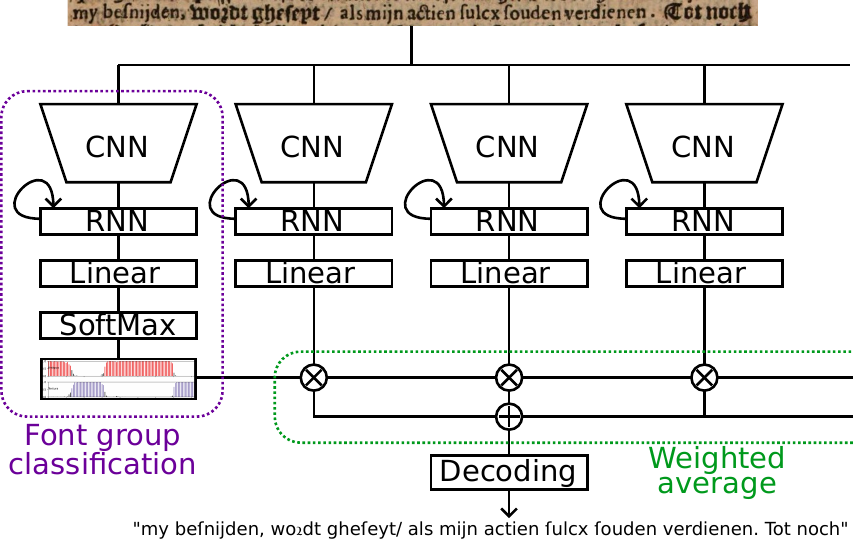}
    \end{center}
    \caption{
        Pipeline of the COCR system.
    }
    \label{fig:cocr}
\end{figure}

We first apply the classifier and all \ac{ocr} models\footnote{We found out that applying only \ac{ocr} models for font groups which have, somewhere in the batch, a score higher than 0.1 improved the speed by roughly 40\,\% with no more than differences of 0.01\,\% of \ac{cer}. However, these values are highly data-dependent.}.
Then, we use the classification scores as weights to make a weighted sum of the \ac{ocr} outputs, before the decoding step.
This is illustrated in \cref{fig:cocr}.

As this method is differentiable, we can train everything together.
However, we started with already-trained networks: the baseline for the different \ac{ocr} networks, and first trained the classifier on its own.
As the \ac{ocr} system gets to see the whole input and must provide outputs that can be combined, we initialize all of them with the weights of the baseline, not with the font-group-specific models.

\subsection{Training Procedure}
The text lines are resized to a height of 32 pixels, and their aspect ratio is preserved.
This height is a good compromise between data size (and \ac{ocr} speed) and readability.
The font group label arrays, which are attributed to every pixel column of the text lines, are scaled accordingly using the nearest value, as interpolating class numbers would be quite meaningless.

The networks are trained with a batch size of 32, for a maximum of 1000 epochs, and we use early stopping with a patience period of 20 epochs.
We selected Adam~\cite{kingma2014adam} as the optimizer, mainly for convergence speed reasons, and the \ac{ctc}~\cite{graves2006connectionist}.
Every 5 epochs without \ac{cer} improvement, the learning rate is halved.
For the baseline, we start with a learning rate of 0.001, and random initial weights.
The other models are initialized with the weights of the baseline, and the optimizer is reloaded as well in order to start the training from the point which had the lowest validation \ac{cer} for the baseline.

The two classifiers, i.e., the one used in the SelOCR and SplitOCR system, and the one used in the COCR system, are trained slightly differently.
In the case of the pixel column classifier, we use a batch size of 1, as this seemed to converge faster and better.
Regarding the classifier used in the COCR system, experiments have shown that if it is not trained before starting to train the system as a whole, results are not competitive.\footnote{No class labels are shown to the classifier when the COCR system is trained so that the classifier has no restriction in how it merges the \ac{ocr} outputs.}
So, in order to train it, as it outputs a sequence shorter than the labels of the pixel columns, we downscaled the pixel-wise ground truth without interpolation.

Offline data augmentation of three different types (warping, morphological operations, and Gaussian filtering) is applied.
As these augmentations are rather fast, applying them at runtime still significantly slowed down the training process. This is why we decided to store the augmented data instead.

Warping is done with ocrodeg;~\cite{githubGitHubNVlabsocrodeg} as parameter we set \texttt{sigma} to 5, and \texttt{maxdelta} to 4.
These values were selected by trying different combinations and retaining one which provided realistic results on some randomly selected images.
As morphological operations, we selected grayscale erosion and dilation with square elements of size $2\times 2$.
For the Gaussian filtering, we have two methods as well: Gaussian blur, with its variance randomly selected in $\left[1, 1.25\right]$, and unsharp filter using the same parameter range.
The verisimilitude of the produced data could likely be enhanced furthermore by adapting all these parameters to the resolution of the images.

\begin{figure}[t]
    \centering
    \begin{subfigure}{1\textwidth}
        \centering
        \includegraphics[width=0.9\textwidth]{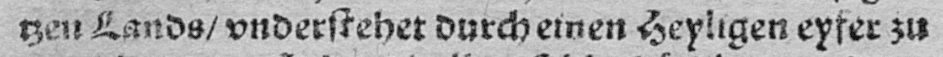}
        \caption{Original}
        \label{fig:aug-original}
    \end{subfigure}
    \begin{subfigure}{1\textwidth}
        \centering
        \includegraphics[width=0.9\textwidth]{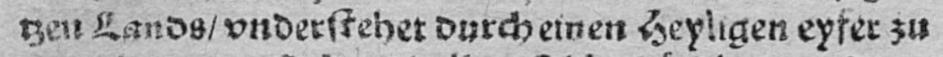}
        \caption{Warping}
        \label{fig:aug-warp}
    \end{subfigure}
    \begin{subfigure}{1\textwidth}
        \centering
        \includegraphics[width=0.9\textwidth]{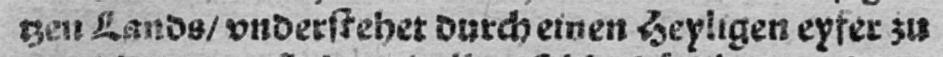}
        \caption{Erosion}
        \label{fig:aug-erosion}
    \end{subfigure}
    \begin{subfigure}{1\textwidth}
        \centering
        \includegraphics[width=0.9\textwidth]{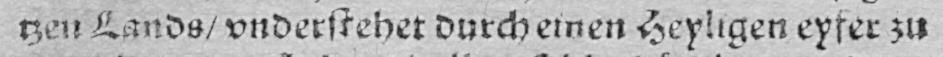}
        \caption{Dilation}
        \label{fig:aug-dilation}
    \end{subfigure}
    \begin{subfigure}{1\textwidth}
        \centering
        \includegraphics[width=0.9\textwidth]{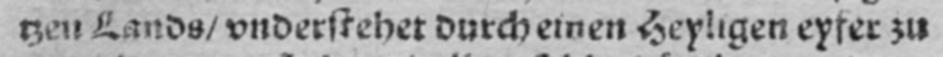}
        \caption{Gaussian}
        \label{fig:aug-gaussian}
    \end{subfigure}
    \begin{subfigure}{1\textwidth}
        \centering
        \includegraphics[width=0.9\textwidth]{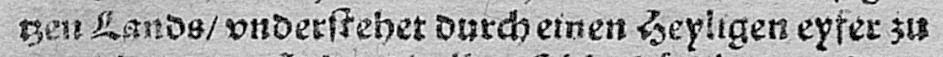}
        \caption{Unsharp}
        \label{fig:aug-unsharp}
    \end{subfigure}
    \caption{Illustration of the offline augmentations which we applied.} 
    \label{fig:augmentations}
\end{figure}

To produce offline augmented images, we selected a random non-empty set of these augmentations, with no more than one augmentation of each kind (i.e., erosion, or dilation, or none of these, but not both erosion and dilation).
For each original training text line, we created two additional ones.
An example of our augmentations is given in \cref{fig:augmentations}.

Moreover, at run time, we use Torchvision~\cite{torchvision2016} to apply further augmentations: shearing (maximum angle of 5 degrees), and modifications of contrast and luminosity (maximum 20\,\%).

\subsection{Evaluation}
Our main metric is the \acf{cer}.
The \ac{cer} between transcription and the corresponding ground truth, is obtained by dividing the Levenshtein distance~\cite{levenshtein1966binary} between these two strings by the length of the ground truth.
In order to avoid averaging together \ac{cer} obtained on short and long text lines, we consider our test set as a whole, and compute the \ac{cer} as the sum of the Levenshtein distances for all text lines, by the sum of the length of all ground truth strings.
We used the \texttt{editdistance} library\footnote{\url{https://pypi.org/project/editdistance/}}, version 0.6.2.
We encode and decode strings using Python iterators, for which reason some special characters are composed of several codepoints.

\section{Data}
\label{sct:data}
Our data consists of 2,506 pages, taken from 849 books printed between the 15th and 18th centuries.
On each of those pages, we have manually drawn bounding boxes around text lines or parts of text lines.
We labeled the (only) font group of each of these boxes~--~thus, if a text line contains more than one font group, then it is split into several bounding boxes. 
We transcribed the content of these bounding boxes.
In total, we annotated over 90,000 bounding boxes and transcribed over 3.6 million characters.
Our data also contains bounding boxes for Greek, Hebrew, and Manuscripts, however as they are a negligible fraction of our data, we omitted them in this study.
More detailed statistics are given in \cref{tab:data}.

Moreover, we considered two super-groups of fonts: Gothic and Roman.
Indeed, as it can happen that Gothic font groups are erroneously considered as \textit{Fraktur} and other ones as \textit{Latin} (e.g., in in~\cite{springmann2018ground,bjerring2022mending}, or the Fraktur and non-Fraktur pre-trained models for the German language of Tesseract\footnote{\url{https://github.com/tesseract-ocr/tessdata\_best}}), we decided to investigate these two subsets of our data.
The first one, Gothic, is composed of Bastarda, Fraktur, Rotunda, Schwabacher, and Textura.
The second one, the Roman font group, is composed of Italic and Antiqua.
We decide not to use the name \textit{Latin}, as some clearly Gothic font groups, such as Textura, are often used for texts in Latin (as a language).
We did not include Gotico-Antiqua, as it is by definition halfway between Roman and Gothic font groups.

Our data is split at the book level so that no data from the same book can be present in both training and test sets.
To produce the test set, we tried a large number of random combinations for which all font groups have at least 10'000 characters, and kept the one minimizing the variance of the number of characters per font group.
Moreover, we manually swapped one Gotico-Antiqua book with another one, as it contained special characters not present in any other book~--~it would make little sense to test the recognition of characters not present in the training data. The validation set was obtained in the same way but contains a smaller amount of data.

\begin{table}[t]
\centering
\caption{Amount of bounding boxes and characters in our dataset.}\label{tab:data}
\begin{tabular}{lcccccc}
    \toprule
        Font group & Training & Validation & Test & Training & Validation & Test \\
    \midrule
        Antiqua & 13725 & \phantom{1}310 & \phantom{2}592 & \phantom{3}400685 & \phantom{4}6166 & \phantom{1}13361 \\
        Bastarda & 13528 & \phantom{12}99 & \phantom{2}301 & \phantom{3}582765 & \phantom{4}5468 & \phantom{1}12088 \\
        Fraktur & 10360 & \phantom{1}163 & \phantom{2}250 & \phantom{3}413798 & \phantom{4}6651 & \phantom{1}12052 \\
        Gotico-Antiqua & \phantom{1}8200 & \phantom{1}143 & \phantom{2}373 & \phantom{3}345171 & \phantom{4}6118 & \phantom{1}12653 \\
        Italic & 11315 & \phantom{1}186 & \phantom{2}443 & \phantom{3}410085 & \phantom{4}6604 & \phantom{1}14645 \\
        Rotunda & \phantom{1}7261 & \phantom{1}132 & \phantom{2}229 & \phantom{3}301323 & \phantom{4}7842 & \phantom{1}14180 \\
        Schwabacher & 15808 & \phantom{12}98 & \phantom{2}432 & \phantom{3}684879 & \phantom{4}5302 & \phantom{1}16182 \\
        Textura & \phantom{1}7724 & \phantom{1}123 & \phantom{2}265 & \phantom{3}338632 & \phantom{4}5714 & \phantom{1}12875 \\
        Total & 87929 & 1254 & 2885 & 3477458 & 49865 & 108036 \\
    \bottomrule
\end{tabular}
\end{table}

\section{Experimental Results}
We evaluate the different approaches following three scenarios.
The first one is simply using all of our test data.
This evaluates how well the different approaches perform if we do not know in advance which font groups are in the data; this is the most realistic use case.
The second one is using only text lines containing multiple font groups, as it allows us to see how the different models deal with font changes.
The last scenario uses the ground truth to separate the font groups.
While the least realistic (it is unlikely an expert would look at each text line to label its font group before passing it through an \ac{ocr} pipeline), but offers insight into the effect of fine-tuning \ac{ocr} models for specific font groups.
We will first briefly present classification results, and expand more on \ac{ocr} results afterward.

\subsection{Classification}
While there is no doubt about which fonts are present in a text line or which font is used for every character, labels for pixel columns are to be considered approximate near class boundaries.
Indeed, the column labels were obtained through manually drawn bounding boxes, thus two experts labeling the same file~--~or the same one labeling it twice~--~would produce slightly different bounding boxes, especially at font group boundaries in white spaces between words.
For this reason, we consider classification accuracy as secondary, as the main task of the classifier is to be included in \ac{ocr} systems.

\begin{figure}[tb]
    \begin{center}
        \includegraphics[width=\textwidth]{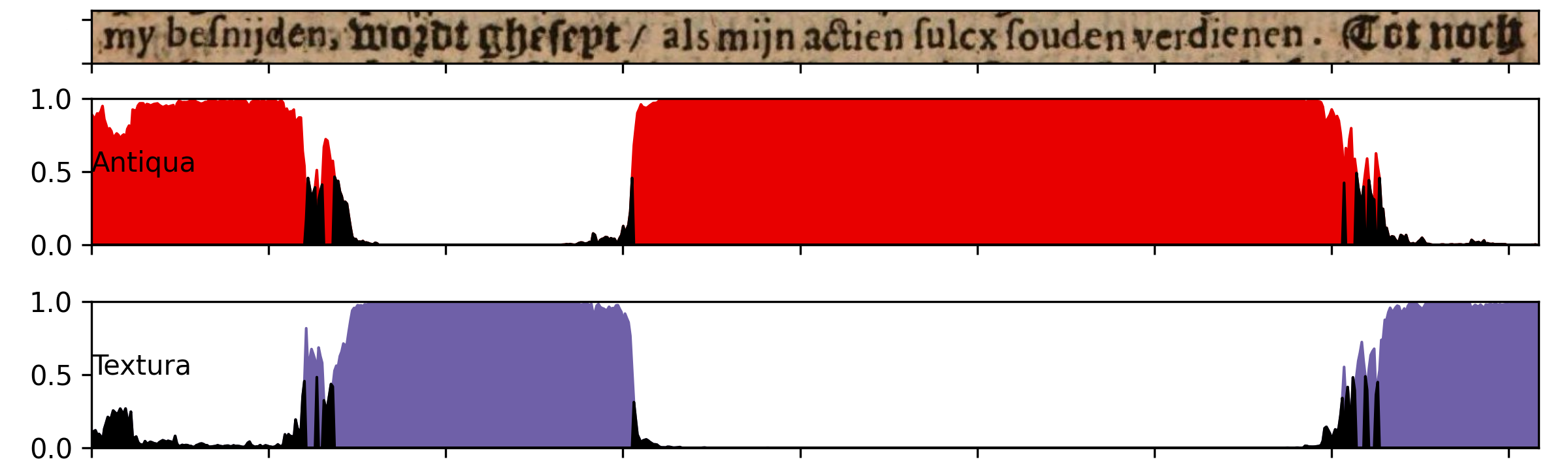}
    \end{center}
    \caption{
        Classification results of pixel columns.
        The two plots correspond to the classification scores for Antiqua and Textura.
        Colors indicate the pixel column with the highest score.
        Results for other font groups are not shown, as they were all extremely close to zero.
    }
    \label{fig:colclassification}
\end{figure}

An example of pixel column classification is shown in \cref{fig:colclassification}.
In this text line, we can see that the two sequences of Antiqua and Textura are mostly correctly identified.
We can, however, notice some fluctuations near class boundaries; this is what we smoothen in the post-processing described in~\cref{ssct:splitocr}.

The classifier used in the COCR system produces an output sequence shorter than the text line, thus we downscale without interpolation the pixels' labels in order to have training data usable by this classifier.

\subsection{OCR}
The results are presented in \cref{tab:results}.
You will notice that the \ac{cer} for the Gotico-Antiqua font group is significantly higher than for other font groups.
As mentioned in \cref{sct:data}, for this font group, we manually selected a document, as the automatic random selection led to having some characters present only in the test data.
We checked the validation \ac{cer} and saw that it matches the ones of other font groups.
Moreover, a test done on another document which is not included in our dataset reached a \ac{cer} only slightly higher than for the other font groups, so we explain the much higher \ac{cer} of Gotico-Antiqua by the difficulty presented by the selected document.
However, for scientific integrity reasons, we did not swap this document with another one and present the results as they are.

First, let us have a look at the baseline and the models with a similar architecture, i.e., the ones which are below it in \cref{tab:results}.
While not being the best, the baseline is performing relatively well on the whole dataset (first column) compared to the other models.
However, for all considered subsets of our data, there is at least one other comparable model with a lower \ac{cer}.
Models fine-tuned on specific font groups always perform better than the baseline for the corresponding font group; this is not a surprise but is nevertheless worth being mentioned.

\begin{table}[t]
\centering
\caption{
    \ac{cer} (\%) for the different \ac{ocr} systems, on the test set and its subsets.
    Systems named as font groups are the ones fine-tuned for this font group.
    The \textit{All} and \textit{Mult.} sets are the whole test set and its text lines which contain multiple font groups, respectively.
    The following subsets are text lines with only the mentioned font group.
    The horizontal and vertical separators help distinguish better results obtained with simple or more complex network architectures, and on subsets with multiple or single font groups, respectively.
}\label{tab:results}
\begin{tabular}{lcccc|cccccccc}
    \toprule
        System & All & Mult. & Got. & Rom. & Ant. & Bas. & Fra. & G.-A. & Ita. & Rot. & Schw. & Tex. \\
    \midrule
        COCR & \textbf{1.81} & \textbf{1.95} & \textbf{1.17} & \textbf{1.89} & 1.76 & 1.36 & 1.05 & 4.44 & \textbf{2.00} & 0.99 & 1.06 & 1.47 \\
        SelOCR & 1.82 & 2.61 & \textbf{1.17} & 1.92 & 1.76 & 1.32 & 1.01 & 4.35 & 2.06 & 1.01 & 1.07 & 1.47 \\
        SplitOCR & 2.19 & 4.83 & 1.48 & 2.19 & 2.11 & 2.02 & 1.59 & 4.65 & 2.26 & 1.07 & 1.22 & 1.64 \\
    \midrule
        Baseline & 1.92 & 2.59 & 1.21 & 2.12 & 1.98 & 1.44 & 1.03 & 4.57 & 2.24 & 1.04 & 1.08 & 1.54 \\
        Gothic & 1.95 & 2.61 & \textbf{1.17} & 2.35 & 2.30 & \textbf{1.26} & \textbf{0.98} & 4.53 & 2.40 & 1.04 & 1.11 & 1.49 \\
        Roman & 2.02 & 2.40 & 1.32 & 2.00 & 1.87 & 1.45 & 1.32 & 5.06 & 2.12 & 1.10 & 1.23 & 1.58 \\
        Antiqua & 1.98 & 2.53 & 1.30 & 1.94 & \textbf{1.75} & 1.44 & 1.20 & 4.93 & 2.11 & 1.16 & 1.17 & 1.57 \\
        Bastarda & 1.97 & 2.67 & 1.24 & 2.17 & 2.09 & 1.31 & 1.05 & 4.67 & 2.25 & 1.14 & 1.11 & 1.63 \\
        Fraktur & 1.95 & 2.67 & 1.21 & 2.10 & 2.05 & 1.39 & \textbf{0.98} & 4.77 & 2.15 & 1.11 & \textbf{1.04} & 1.59 \\
        Gotico-Antiqua & 1.90 & 2.53 & 1.23 & 2.16 & 2.14 & 1.38 & 1.09 & \textbf{4.30} & 2.18 & 1.06 & 1.13 & 1.54 \\
        Gotico-Antiqua$^+$ & 1.91 & 2.56 & 1.23 & 2.07 & 1.82 & 1.42 & 1.10 & 4.50 & 2.29 & \textbf{0.96} & 1.15 & 1.60 \\
        Italic & 1.90 & 2.38 & 1.22 & 1.93 & 1.78 & 1.43 & \textbf{0.98} & 4.81 & 2.06 & 1.01 & 1.13 & 1.57 \\
        Rotunda & 1.95 & 2.53 & 1.22 & 2.16 & 2.12 & 1.40 & 1.00 & 4.69 & 2.19 & 1.01 & 1.12 & 1.60 \\
        Schwabacher & 1.97 & 2.59 & 1.19 & 2.24 & 2.16 & 1.34 & 1.03 & 4.87 & 2.31 & \textbf{0.96} & 1.06 & 1.61 \\
        Textura & 1.94 & 2.53 & 1.20 & 2.10 & 1.85 & 1.42 & 1.10 & 4.81 & 2.32 & 0.99 & 1.07 & \textbf{1.45} \\
        Best Fine-Tuned & 1.90 & 2.40 & \textbf{1.17} & 1.93 & \textbf{1.75} & \textbf{1.26} & \textbf{0.98} & \textbf{4.30} & 2.06 & \textbf{0.96} & \textbf{1.04} & \textbf{1.45} \\
    \bottomrule
\end{tabular}
\end{table}

We can also see that results for the fine-tuned models get the best results for their architecture six times out of eight, with the exception of the ones for Schwabacher and Rotunda.
Moreover, only one of the complex models performs best on one single font (Italic).

Out of curiosity, we also trained an extra model for Gotico-Antiqua, which we named Gotico-Antiqua$^+$.
As this font group is by definition halfway between Gothic and Roman font groups, we added training data from the visually most similar font groups, namely Antiqua and Rotunda.
While this improves its \ac{cer} on these two other groups, it also has a negative impact on Gotico-Antiqua.

Now, let us focus on the three other systems, which are based on font group identification.
The SplitOCR system has the highest \ac{cer} of all methods evaluated in this paper.
This is due to the fact that when text lines are split into smaller parts~--~more specifically, we think that shorter text lines do not provide enough contextual information for the \ac{rnn} layers to perform well.
Moreover, another drawback of this method is that text lines have to be processed one by one, and multiple calls to \ac{ocr} models, one per segment of the text line, have to be made.

\begin{figure}[tb]
    \begin{center}
        \includegraphics[width=\textwidth]{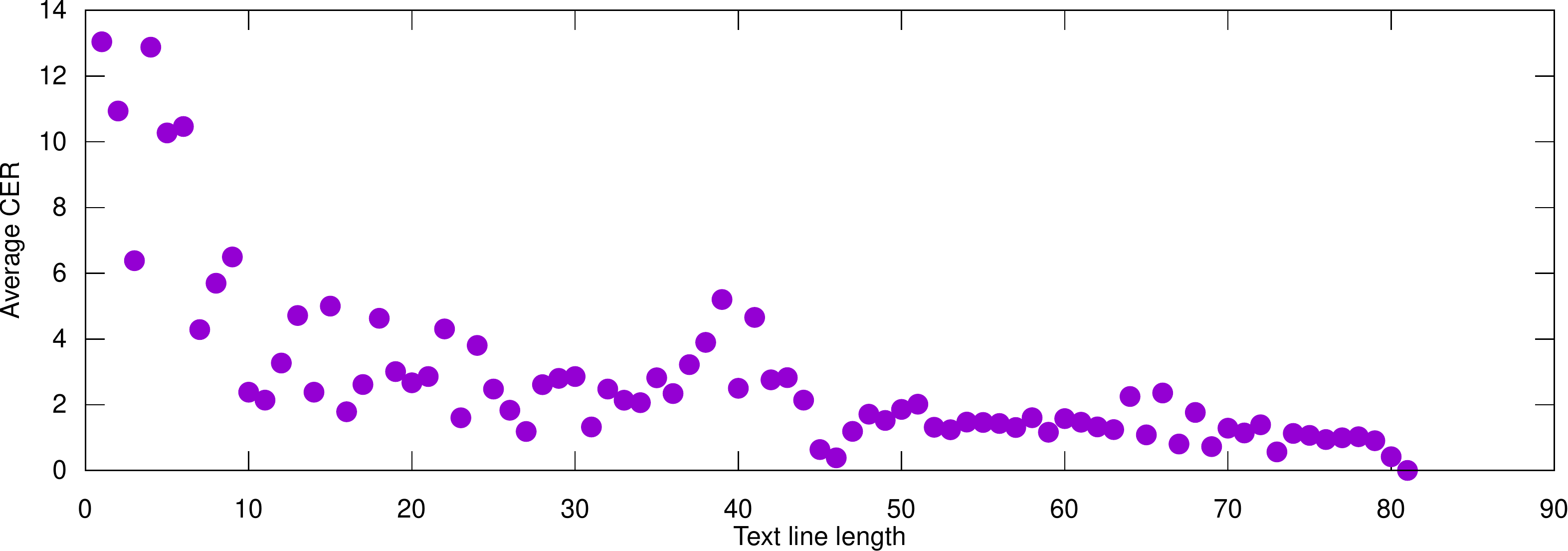}
    \end{center}
    \caption{
        Mean \ac{cer} for every text line length in our dataset.
    }
    \label{fig:linelength}
\end{figure}

In order to investigate this, we computed the baseline \ac{cer} of all test lines of our dataset, grouped the lines in bins based on their number of characters, and computed the average \ac{cer} of each bin.
The plot we produced is shown in \cref{fig:linelength}.
We can make two main observations.
First, shorter text lines tend to have a higher \ac{cer}, which supports our previous assumption.
Second, the variance of the results is higher for shorter text lines.
This is not caused by a lower amount of short lines (there are almost 500 of length 1 to 10, and 300 of length 71 to 81), but rather by the higher \ac{cer} introduced by errors in shorter lines.

The SelOCR system works significantly better than the baseline, with a \ac{cer} 9.4\,\% lower.
It uses the \ac{ocr} models fine-tuned for the different font groups, and thus differences in \ac{cer} are entirely due to the results of the classifier which selects which \ac{ocr} model to apply.
At the cost of first classifying text lines, it beats the baseline, as could be expected, in all cases, except for the subset of text lines containing multiple font groups.
Also, at least in our implementation, text lines have to be processed one by one, as a condition is used for applying the right \ac{ocr} model.

The COCR system is slightly better than the SelOCR on the whole dataset.
We, however, think that such a small difference is not significant; with a little different test set, the tendency could reverse.
For this reason, we would not consider one being better than the other on the whole set.
However, if we look at the second column, i.\,e., text lines containing multiple fonts, we can see that the COCR system does perform significantly better, with a \ac{cer} a quarter smaller.

\section{Discussion \& Conclusion}
In the previous sections, we presented and investigated various \ac{ocr} methods well suited for early modern printed books.
We showed that font groups have a significant impact on \ac{ocr} results, and proved the advantage of fine-tuning models on a granularity finer than Roman and Gothic font groups~--~this highlights the need to identify these font groups.

Also, we developed and presented a novel method, the COCR system, in which performances are only slightly impacted by the presence of multiple font groups in a text line.
However, it is computationally more expensive than the SelOCR, which applies only one \ac{ocr} model after identifying the main font of a text line.
Considering they reach almost identical \ac{cer} on our data, we would recommend using this approach only for documents that have frequent font group changes inside of text lines.

Now that we showed the potential of combining font group identification and \ac{ocr}, our future work will consist in investigating this topic further.
We are especially interested in knowing if some parts of the \ac{ocr} models of the COCR system, mentioned in \cref{tab:architecture}, can be merged in order to save memory, and process data faster~--~and be environmentally more friendly by saving energy.
Moreover, we would like to examine in detail the outputs of the classifier used in the COCR system, as it surprisingly seems not to perform well for classification once the system has been fine-tuned for \ac{ocr}.

\bibliographystyle{splncs04}
\bibliography{bibliography}
\end{document}